# EmoBang: Detecting Emotion From Bengali Texts

Abdullah Al Maruf[1], Aditi Golder[2], Zakaria Masud Jiyad[1], Abdullah Al Numan[3], and Tarannum Shaila Zaman[1][4]

[1] Charles Darwin University, Darwin, Australia
{s395350, s393850}@students.cdu.edu.au
[2] Wake Forest University, Winston-Salem, NC 27109, United States golda24@wfu.edu
[3] Bangladesh University of Engineering and Technology (BUET), Dhaka, Bangladesh
numan290.aan@gmail.com
[4] University of Maryland, Baltimore County (UMBC), Maryland, United States
zamant@umbc.edu

**Abstract** Emotion detection from text seeks to identify an individual's emotional or mental state, positive, negative, or neutral, based on linguistic cues. While significant progress has been made for English and other high-resource languages, Bengali remains underexplored despite being the world's fourth most spoken language. The lack of large, standardized datasets classifies Bengali as a low-resource language for emotion detection. Existing studies mainly employ classical machine learning models with traditional feature engineering, yielding limited performance. In this paper, we introduce a new Bengali emotion dataset annotated across eight emotion categories and proposes two models for automatic emotion detection: (i) a hybrid Convolutional Recurrent Neural Network

(CRNN) model ($EmoBang_{Hybrid}$) and (ii) an AdaBoost–Bidirectional Encoder Representations from Transformers (BERT) ensemble model ($EmoBang_{Ensemble}$). Additionally, we evaluate six baseline models with five feature engineering techniques and assess zero-shot and few-shot large language models (LLMs) on the dataset. To the best of our knowledge, this is the first comprehensive benchmark for Bengali emotion detection. Experimental results show that $EmoBang_H$ and $EmoBang_E$ achieve accuracies of 92.86% and 93.69%, respectively, outperforming existing methods and establishing strong baselines for future research.

## 1 Introduction

Human emotion may be expressed in various ways, including verbal, textual, and facial expressions. As emotion plays a vital role in decision-making [10], it can influence marketing, business, politics, human behavior, technology, and other essential aspects of our lives. Due to its pervasive application in human life, emotion detection from texts has become increasingly crucial [39]. Emotion detection from texts aims to identify the writers' hidden feelings in their written works. The hypothesis of this technique is based on the assumption that if an author is happy, they use positive words. If an author is sad or frustrated, they use negative words [10].

---

[1] Corresponding author

According to Natural language processing (NLP), emotion detection is identifying and classifying emotions in written texts [42]. Websites, social media platforms, news sites, emails, online stores, user feedback services, customer service inquiries, etc., are all frequent sources of unstructured text data [19] used by the researchers. These sources can supply a significant volume of data, but the data might be challenging because of their unorganized nature. Researchers have made notable progress in detecting emotions in high-resource languages such as English, French, and Chinese [7]. However, the field of recognizing emotions through textual data in low-resource languages is still in a growing stage and attracting considerable attention [7,41].

Bengali is considered a low-resource language in NLP due to the limited availability of annotated data and resources for Bangla language processing [31]. Bengali, with approximately 250 million native speakers worldwide, ranks fourth among the most widely ence of Bespoken languages [12]. However, emotion detection in Bengali is still an under-explored area [16]. The limitation of annotated data and cultural variability in emotions make emotion detection from Bengali texts more challenging. However, it has significant applications for customer service, marketing and advertising, human resource management, healthcare, and security.

Several existing studies have focused on emotion detection from Bengali text [35,6,20]. However, most of these works rely on limited datasets [35,6], resulting in relatively low accuracy. For instance, traditional machine learning approaches such as Support Vector Machine (SVM) [35] and Naïve Bayes (NB) [6] achieved accuracies of 73% and 78%, respectively, in Bengali emotion classification tasks. Some recent methods have reported improved results, such as the approach proposed by Vijay et al. [20], which employs a manually annotated dataset and a pre-trained model. Their hybrid architecture—combining a Word2Vec embedding layer with Convolutional Neural Network (CNN) and Long ShortTerm Memory (LSTM) networks—achieved an accuracy of 90.49%. However, this model is restricted to detecting only three basic emotions: happiness, anger, and sadness. Furthermore, Large Language Model (LLM)–based approaches [29] for Bengali emotion detection require further exploration.

To address these challenges, in this paper we propose two novel techniques, *EmoBang$_{Hybrid}$* and *EmoBang$_{Ensemble}$*, for detecting eight distinct emotions from Bengali texts. The first approach, *EmoBang$_{Hybrid}$*, employs a Convolutional Recurrent Neural Network–Long Short-Term Memory (CRNN-LSTM) based hybrid architecture, while the second, *EmoBang$_{Ensemble}$*, introduces an ensemble framework that, to the best of our knowledge, has not been previously applied to Bengali emotion detection tasks. Both proposed methods demonstrate superior performance compared to existing approaches. Furthermore, this study provides a comprehensive empirical analysis by applying six traditional machine learning (baseline) models and two Large Language Models (LLMs) to the proposed dataset. The findings of this study are expected to serve as a valuable benchmark and foundation for future research on emotion detection in Bengali texts.

We apply five feature engineering techniques and conduct experiments, selecting BERT [13] for our models. We propose two models, $EmoBang_{ensemble}$ and $EmoBang_{hybrid}$, to evaluate performance. The hybrid model combines CRNN and LSTM, while the ensemble model uses an Adaboost classifier. Using multiple evaluation metrics, we find both models perform best with BERT features, with the ensemble slightly outperforming the hybrid. Section 3 provides detailed model descriptions. We address five research questions: i) RQ1: How effectively do EmoBang models detect emotions from Bengali texts? ii) RQ2: How does data balancing affect model performance? iii) RQ3: How do feature engineering techniques improve emotion detection? iv) RQ4: Does $EmoBang_{Hybrid}$ outperform the existing hybrid model [20]? v) RQ5: How do LLMs perform on our dataset? In summary, in this paper, we make the following contributions:

– We collect a Bengali real-world dataset from four sources and classify it into eight emotional classes.
– We conduct a comprehensive study using six baseline classifiers and five feature engineering techniques, and we also evaluate two LLM models. This multi-faceted analysis reveals key behavioral patterns in Bengali text data.
– We propose a novel hybrid model, $EmoBang_{Hybrid}$, based on CRNN, which achieves precision 92.67, recall 91.15, F1-score 92.50, and accuracy 92.86.
– We develop a BERT-based ensemble model, $EmoBang_{ensemble}$, the first of its kind for Bengali texts, which achieves precision 93.43, recall 92.43, F1score 93.67, and accuracy 93.69.

## 2 Related Work

**Emotion Detection from Bengali Texts:** There are many existing works [32,30,35,9,4,17,28,34,44,45] where researchers detect emotions from Bengali texts. The summary of these works shows that in most of the works [32,35,9,4,28], the number of emotion classes used for labelling is less than our proposed work EmoBang. Existing works can detect a maximum of six different emotion classes [45,44,32] and EmoBang can detect eight. For example, Rahman et al. [32] can classify six emotion classes, and the accuracy of their approach is 52.98%.

Some of the works [4,30] developed to detect emotion from Bengali texts do not provide detailed size and nature of the dataset. Therefore, no further study can be done on those datasets. Many existing techniques[9,35] use relatively small datasets which may not show reliable results. For example, Bhowmik et al. [9] collect a small training ABSA dataset and focus on only the feature words, which may limit the generalizability of the work. Ruposh et al. [35] use a relatively small corpus of 1200 emotive words for training the SVM classifier and do not provide detailed information about the data pre-processing techniques. Compared to these works, our EmoBang contains a standard set of data (4240 Bengali sentences). We will share our dataset publicly so that further research can be done. We also preprocess and balance our data, a novel feature in detecting emotion from Bengali texts.
**Emotion detection from Bengali text using LLM:** Despite advances in multilingual NLP, Large Language Models (LLMs) remain underused in Bangla for sentiment and emotion analysis. Sadhu et al. [36] show that LLMs reproduce sociocultural biases, linking men with authority or anger and women with empathy or fear, highlighting the need for fairness-aware modeling in low-resource languages. Tabassum [43] analyzes Bangladeshi e-commerce reviews with BERT, mBERT, and LLaMA, finding that optimized LLaMA-3.1-8B achieves 95.5% accuracy while remaining computationally efficient via LoRA and PEFT. These studies highlight both the promise and ethical challenges of LLMs in Bangla, emphasizing bias detection, interpretability, and efficient adaptation.

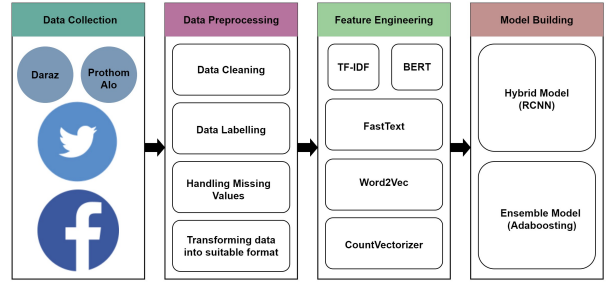

Figure1: Framework of the EmoBang

## 3   Methodology

Figure 1 depicts the overview of EmoBang. It has four major steps: i) Data Collection, ii) Data Preprocessing, iii) Feature Engineering, and iv) Model Building. In this study, we have worked with eight emotion types from Bengali texts:
Anger, Sadness, Happiness, Disgust, Sarcastic, Fear, Surprise, and Disappointed.

### 3.1   Data Collection

We have gathered a total of 4240 Bengali sentences and approximately 44650 words. While there are several available datasets [40,21,44,6] for emotion detection, many of them have limited amounts of data[6,21,44], and researchers use a maximum of six classes[6,27] for conventional datasets. To compare and contrast our findings with these existing datasets, we have created our dataset that includes eight distinct classes (Anger, Sadness, Happiness, Disgust, Sarcastic, Fear, Surprise, and Disappointed). For detecting emotions from Bengali texts, the data is collected from various sources, including popular social media platforms such as Twitter and Facebook, as well as popular news websites such as Prothom Alo [2] and e-commerce sites Daraz [1]. We use API [11] techniques based on different contents to collect data from Facebook and Twitter. We use scrapping [14] to collect data from the popular newspaper website Prothom Alo and the E-Commerce site Daraz, where thousands of users post their feedback in the form of comments and posts daily.

## 3.2 Data Preprocessing and Labelling

Text preprocessing in Bengali follows the same general principles as other languages but presents unique challenges. We first remove special characters, diacritics, punctuation marks, HTML tags, URLs, and digits from the text to reduce noise and improve model accuracy.

We convert emojis to their corresponding Bengali emotions to better capture sentiment. Using Python's "emoji" library, we map emojis to their Unicode representations and then to Bengali emotion words. For example, 😃 emoji, representing happiness, maps to ``আনন্দময়'' (Anondomoy). Some emojis have different meanings in Bengali than in English, so we perform manual conversions. The "folded hands" emoji, for instance, often expresses gratitude or a request for forgiveness in Bengali, whereas in English it usually indicates praying or a high-five.

Removing stop words is challenging in Bengali due to the lack of dedicated packages, so we manually create a stop-word list (Table 1). We apply the Syn-thetic Minority Oversampling Technique (SMOTE) [26] to balance the dataset by oversampling the minority class. SMOTE identifies instances in the underrepresented class, randomly selects one, and finds its k-nearest neighbors, where k is a user-defined parameter. It generates synthetic samples by interpolating feature values between the selected instance and its neighbors. We repeat this process until reaching the desired oversampling level, which balances the class distribution, reduces overfitting, and improves model generalization [23].

Table 1: Created stop words functions that remove the listed stop words

| SW(Bng) | Eng. | SW(Bng) | Eng. | SW(Bng) | Eng. | SW(Bng) | Eng. |
|---|---|---|---|---|---|---|---|
| এ | this | যায় | possible | এর | of this | হয় | is/are |
| কি | what | বা | or | যাক | let's go | য | who/that |
| ক | who | সব | all | উপর | on/above | হব | will be |
| এই | this | একই | same | তাকে | him/her | আগে | before |
| বা | or | কখন | when | আছ | has/have | তাই | therefore |
| স | this | সই | that | হয় | happens | তার | his/her/its |
| যিদ | if | অধীন | under | কের | does/makes | ছিল | was/were |
| আিম | I | এবং | and | তারা | they | কার | whose |
| িট | this | গুিল | plural | হত | to be | সটা | that |
| আরও | more | খবু | very | পের | after | কান | which |
| কন | why | সকল | all | িঠক | correct | যারা | those who |
| কান | any | তুিম | you | | | | |

We label our data manually, following common practice in Bengali emotion, sentiment, and hate speech detection [6,9,25]. Three student volunteers and a supervisor, who has strong command of Bengali, assign emotion labels to each sentence. We use a voting technique [33]: each participant labels every sentence, and we assign the emotion class with the majority of votes.

### 3.3   Feature Engineering

Feature extraction, or vectorization, converts raw text into numerical features that serve as inputs for machine learning models [37]. This process encodes words in a text document as fixed-length vectors that capture their semantic meaning. The choice of feature extraction method strongly affects the performance of a text classifier, making it essential to experiment with different techniques for each task. In our study, we apply both classical and modern feature engineering techniques, including CountVectorizer [5], TF-IDF [18], BERT [13], Word2Vec [24], and FastText [22], to extract features from the text data. Our experiments show that BERT embeddings [38] perform best. Therefore, we use BERT embeddings in our proposed ensemble and hybrid models to enhance performance. We also fine-tune the BERT model to generate task-specific embeddings, further improving our results.

### 3.4   Proposed Model

We propose two models: (i) the hybrid model ($EmoBang_{hybrid}$) and (ii) the ensemble model ($EmoBang_{ensemble}$). While many studies use hybrid models for emotion detection, we introduce a new hybrid model that outperforms previous approaches[20]. We also present an ensemble-based model, which, to our knowledge, no one has yet applied to Bengali text emotion detection.

We propose a novel AdaBoost-BERT ensemble model, $EmoBang_{ensemble}$, for multiclass emotion detection in Bengali text. The model enhances AdaBoost by integrating a fine-tuned BERT transformer, inspired by Gao et al.'s work on legal text analysis [15]. $EmoBang_{ensemble}$ captures latent textual features by reweighting weak classifiers and leveraging BERT's output probabilities. BERT performs internal tokenization, converting raw text into token IDs, which pass through transformer layers to produce contextualized embeddings.

We refine these embeddings using additional neural layers: a dropout layer prevents overfitting, and a dense layer with eight units maps embeddings to the emotion classes. Softmax converts the logits into a probability distribution. AdaBoost then forms an ensemble of weak classifiers, adjusting their weights based on performance and emphasizing those that perform well. This step combines

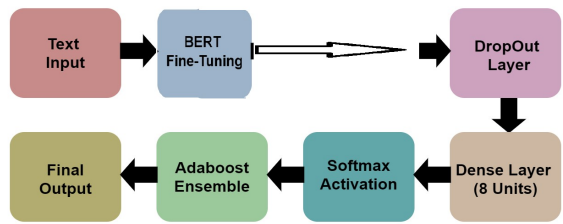

Figure2: Adaboost model architecture

the strengths of multiple BERT-based classifiers, further improving classification accuracy.

The AdaBoost-BERT model is trained on a set of Bengali texts and their corresponding emotion labels, denoted as $X = \{(x_1,y_1),...,(x_m,y_m)\}$. At each iteration $t$, the model computes the classifier $h_t(x)$ that minimizes the weighted error $E_t$:

$$E_t = \sum_{i=1}^{m} w_{t,i} \cdot I(h_t(x_i) \neq y_i)$$

Here, $w_{t,i}$ represents the weight of the $i$-th sample at iteration $t$, and the indicator function $I(\cdot)$ is used to assess misclassification. The weights are updated for accurately classified samples as follows: $w_{t+1,i} = w_{t,i} \cdot e^{-\alpha_t Z_t h_t(x_i) y_i}$ [?]

In this equation, $\alpha_t$ denotes the weight of the $t$-th weak classifier, and $Z_t$ is a normalization constant ensuring the weights sum to one. The final prediction for a new input $x$ is given by: $H(x) = \mathrm{argmax}_j \sum_{t=1}^{T} \alpha_t I(h_t(x) = j)$

This approach integrates BERT's deep contextual understanding with the AdaBoost algorithm, which focuses on combining multiple weak classifiers to optimize the model's performance for multiclass emotion detection from Bengali texts. The AdaBoost-BERT ensemble model leverages the individual strengths of BERT's pre-trained transformer-based neural network and the AdaBoost algorithm's ability to create an ensemble of weak classifiers, achieving improved accuracy in emotion classification tasks.

The BERT model was fine-tuned using the Adam optimizer, with a learning rate of 2e-5 and a batch size of 32. The training process spanned 100 epochs, with early stopping based on the validation loss. A dropout layer with a rate of 0.1 was applied after the BERT output, followed by a dense layer with 8 units and a softmax activation. The softmax function converts the neural network's output into a probability distribution over the possible classes, defined mathematically as:

$$softmax(z_i) = \frac{e^{z_i}}{\sum_{j=1}^{K} e^{z_j}}$$

Here, $z_i$ is the input to the $i^{th}$ neuron in a neural network layer, representing the log-odds or logit of the probability that the input belongs to the $i^{th}$ class.

Table 2: Performance and Execution Time

| Mdl. | Prec. | Rec. | F1 | Acc. | T(m) |
|---|---|---|---|---|---|
| H | 92.67 | 91.15 | 92.50 | 92.86 | 5.6 |
| Ens | 93.43 | 92.43 | 93.67 | 93.69 | 6 |

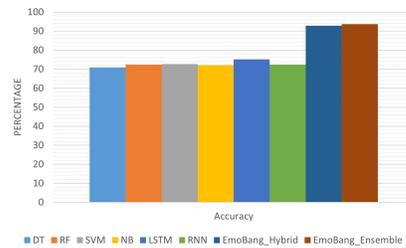

Figure 3: Accuracy Comparison Among Multiple Models

The softmax function returns a probability distribution over the $K$ possible classes, where $K$ is the number of neurons in the output layer.

Overall, the AdaBoost-BERT ensemble model combines the strengths of BERT's pre-trained transformer-based neural network and the AdaBoost algorithm's ability to create an ensemble of weak classifiers to achieve improved performance in multiclass emotion detection from Bengali texts.

## 4   Results and Analysis

**RQ1: Effectiveness and Efficiency of EmoBang Models.** To measure the effectiveness of EmoBang methods we evaluate precision (Prec.), recall (Rec.), F1 score (F1), and accuracy (Acc.) of it. Table 2 shows the performance of $EmoBang_{Hybrid}$ and $EmoBang_{Ensemble}$. The results show that both of the methods have accuracy rates of 92.86% and 93.69% respectively. The $Ensemble$ method has 0.76%, 1.28%, 1.17%, and 0.83% higher precision, recall, F1 score and accuracy respectively compared to $EmoBang_{Hybrid}$. However, both models have more than 90% value in all four measuring parameters. Figure 3 compares accuracy among EmoBang models with the baseline models. This chart shows that $EmoBang_{Hybrid}$ and $EmoBang_{Ensemble}$ models have more accuracy than any other baseline models. $EmoBang_{Hybrid}$ is 22%, 20.41%, 20.24%, 20.74%, 17.65%, and 20.44% more accurate than the baseline models DT, RF, SVM, NB, LSTM, and RNN respectively. On average $EmoBang_{Hybrid}$ is 20.65% more accurate in detecting emotion from Bengali texts than any of the six baseline models. $EmoBang_{Ensemble}$ is 22.83%, 21.24%, 21.07%, 21.57%, 18.48%, and 21.27% more accurate than DT, RF, SVM, NB, LSTM, and RNN models respectively. On average $EmoBang_{Ensemble}$ is 21.08% more accurate than the baseline models. Therefore, EmoBang methods are effective in emotion detection for Bengali texts.

The last column of table 2 shows the total execution time $EmoBang_{Hybrid}$ and $EmoBang_{Ensemble}$ take to detect emotion from Bengali text for our given dataset. This result indicates that EmoBang is efficient enough to detect emotion in Bengali text.

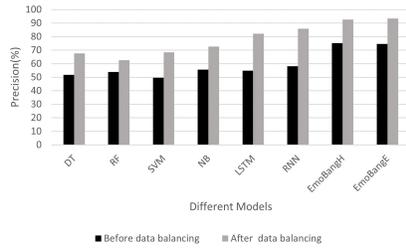 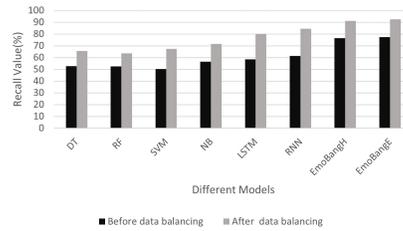

Figure4: Comparison of Precision Value      Figure5: Comparison of Recall Value

**RQ2: Role of Data Balancing in the Effectiveness of Baseline Models and EmoBang Models.** In this work, we perform data balancing and experimental results show that it improves not only the effectiveness of EmoBang models but also all other baseline models. Figure 4 shows that the precision value increases after data balancing for all the six baseline models and for two EmoBang models. After data balancing, the precision value increases by 16.03%,
8.67%, 18.87%, 16.96%, 27.39%, 27.56%, 17.44%, and 18.84% in the DT, RF,

SVM, NB, LSTM, RNN, $EmoBang_{Hybrid}$, and $EmoBang_{Ensemble}$ model respectively. On average it increases by 20.73%. Figure 5 compares the recall values for eight models before and after data balancing. For all of the eight models the recall value increases after data balancing. The experimental evaluation shows that it increases by 13.02%, 11.11%, 17.05%, 15.14%, 21.55%, 23.22%, 14.68%, and 15.11% in DT, RF, SVM, NB,

LSTM, RNN, $EmoBang_{Hybrid}$, and $EmoBang_{Ensemble}$ model respectively. On average the recall value increases by 16.36% after data balancing. Figure 6 shows the comparison among eight differ-

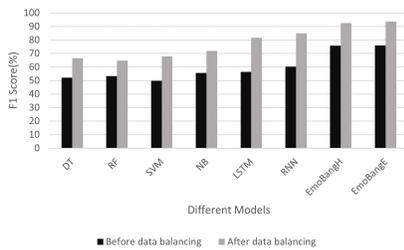

Figure6: Comparison of F1-Score

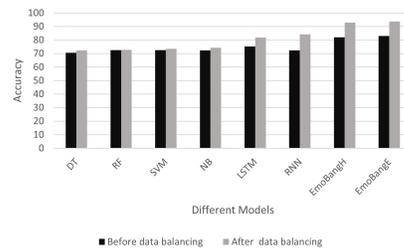

Figure7: Comparison of Accuracy Value

ent models' F1-score before and after data balancing. F1 score is increased by 14.33%, 11.46%, 18.01%, 16.24%, 25.18%, 24.64%, 16.65%, and 17.71% for DT, RF, SVM, NB, LSTM, RNN, $EmoBang_{Hybrid}$, and $EmoBang_{Ensemble}$ model respectively after balancing the data. On average, the F1 score is increased by 18.03%. Figure 7 shows that the accuracy gets improved in all models after data balancing. Particularly the accuracy is improved by 1.61%, 0.34%, 0.92%, 2.05%, 6.71%, 11.79%, 10.86%, and 10.69% respectively. On average, the accuracy is increased by 5.62% after balancing the data.

Table 3: Performance of Different Feature Engineering Techniques in Different Models

| Models | Metric | CountVectorizer | TF-IDF | Word2Vec | FastText | BERT |
|---|---|---|---|---|---|---|
| DT | Precision | 65.66 | 70.03 | 75.63 | 77.44 | 67.72 |
|  | Recall | 62.3 | 71.9 | 74.87 | 74.88 | 65.66 |
|  | F1-score | 66.10 | 73.3 | 75.86 | 75.50 | 66.49 |
| RF | Precision | 62.2 | 66.87 | 77.8 | 73.8 | 62.54 |
|  | Recall | 63.7 | 65.98 | 71.10 | 72.10 | 63.65 |
| | F1-score | 64.01 | 65.50 | 77.98 | 70.54 | 64.69 |
| SVM | Precision | 67.11 | 73.25 | 80.87 | 77.49 | 68.41 |
| | Recall | 66.8 | 72.10 | 79.32 | 76.32 | 67.23 |
| | F1-score | 67.7 | 73.23 | 80.10 | 77.50 | 67.87 |
| NB | Precision | 58.89 | 64.87 | 72.30 | 72.30 | 72.65 |
| | Recall | 57.41 | 62.76 | 71.55 | 71.55 | 71.55 |
| | F1-score | 58.90 | 62.10 | 71.93 | 71.93 | 71.93 |
| LSTM | Precision | 70.24 | 74.25 | 82.32 | 80.55 | 82.15 |
| | Recall | 68.72 | 73.54 | 81.54 | 79.43 | 80.09 |
| | F1-score | 70.29 | 74.90 | 82.20 | 82.42 | 81.61 |
| RNN | Precision | 67.20 | 71.23 | 83.30 | 83.89 | 85.76 |
| | Recall | 65.32 | 70.23 | 82.43 | 82.54 | 84.54 |
| | F1-score | 66.34 | 71.22 | 82.50 | 82.50 | 84.98 |
| $EmoBang_H$ | Precision | 80.13 | 82.54 | 90.15 | 88.76 | 92.67 |
| | Recall | 79.54 | 81.45 | 89.10 | 87.45 | 91.15 |
| | F1-score | 66.49 | 82.75 | 89.12 | 88.10 | 92.50 |
| $EmoBang_E$ | Precision | 84.51 | 81.21 | 87.65 | 90.54 | 93.43 |
| | Recall | 82.62 | 83.64 | 86.23 | 89.65 | 92.43 |
| | F1-score | 83.71 | 82.32 | 86.99 | 90.05 | 93.67 |

**RQ3: Role of feature engineering techniques for emotion detection from Bengali texts.** We evaluated five different feature engineering techniques:
i) CountVectorizer, ii) Tf-Idf, iii) Word2vec, iv) FastText, and v) BERT by incorporating them into eight different models. Table 3 represents the precision, recall, and F1 score for five different feature engineering techniques on eight different models. These results indicate that the feature engineering technique BERT outperforms all other techniques. Hence, the FastText also performs well, achieving high F1 score values with both proposed models. Word2vec performs comparatively low compared to the other feature engineering techniques.

Table 3 represents the precision, recall, and F1 score for five different feature engineering techniques on eight different classifiers. The findings suggest that the CountVectorizer demonstrates the lowest precision, recall, and F1 score value when combined with various models but exhibits the highest value when employed in the $EmoBang_E$ model. The DT classifier exhibits the most noteworthy precision and recall values when utilizing the FastText approach. The F1 score is also maximized when employing the Word2Vec feature engineering technique. RF produces the lowest recall value employing BERT and the highest precision and F1 score with Word2Vec. In addition, Word2Vec feature engineering obtained the highest precision, recall, and F1 score in conjunction with the SVM classifier. The TF-IDF shows the highest precision and F1 score while utilizing the $EmoBang_E$ model. The NB model achieves the highest precision score in BERT and maximized recall and F1 score while employing Word2Vec, FastText, and BERT feature engineering techniques. The FastText feature engineering technique produces the highest precision, recall, and F1 scores in $EmoBang_E$. Moreover, RNN achieved the best precision, recall, and F1 scores in the BERT technique. Our proposed method exhibits the highest precision, recall, and F1 score value when using the BERT technique. The $EmoBang_H$ model produced the lowest results in all three evaluation matrices while the $EmoBang_E$ model achieved the lowest precision and F1 score utilizing the TF-IDF technique. In summary, different feature engineering techniques performed best for the different classifiers. In the context of the DT classifier, FastText performs best among all other feature engineering techniques.

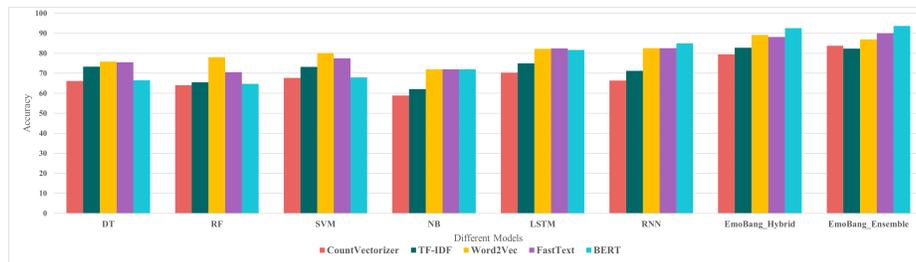

Figure8: Accuracy Comparison of Multiple Models Using Feature Engineering

Figure 8 shows the accuracy comparison for eight different models with five different feature engineering techniques. In the case of the DT, RF, and SVM models, the Word2vec feature engineering techniques show the highest accuracy and CountVectorizer shows the lowest accuracy because words' semantic relationships can be captured using word2vec. The LSTM model shows high accuracy(82.42%) with the FastText feature engineering. Other models, NB, RNN, $EmoBang_{Hybrid}$, and $EmoBang_{Ensemble}$ show the highest accuracy by using the BERT technique. Hence, In our proposed models we use BERT [13] feature Engineering.

**RQ4: Comparison of EmoBang Hybrid Model with the Existing Hybrid Model.** To the best of our knowledge there is an existing hybrid model [20] developed by Muntasir et al. to extract emotion from Bangla texts. No existing techniques use the Adaboost ensemble model to perform this task. Hence we, compare $EmoBang_{Hybrid}$ with the existing hybrid model [20]. Figure 9 shows the comparison between the existing hybrid model [20] and $EmoBang_{Hybrid}$. In the existing hybrid model, authors used three different feature engineering tech-

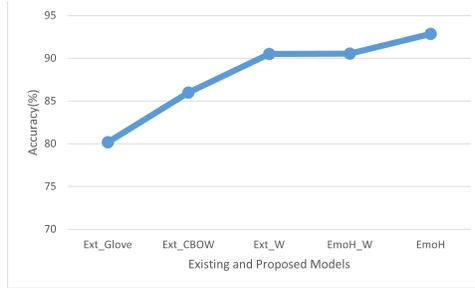

Figure9: Accuracy Comparison of Hybrid Models

niques Glove, CBOW, Word2Vec which is denoted as $Ext\_Glove$, $Ext\_CBOW$, and $Ext\_Word2Vec$ in figure 9. We compare the accuracy of these three variants of the existing hybrid model with our proposed hybrid model $EmoBang_{Hybrid}$ (denoted as EmoH in figure 9). The results show that $EmoBang_{Hybrid}$ has higher accuracy than all of the three variants of the existing hybrid model. In particular, it has 12.7%, 6.86%, and 2.37% higher accuracy than the existing hybrid model with Glove, CBOW, and Word2Vec feature engineering techniques, respectively. Moreover, we also use the Word2Vec feature engineering technique with our proposed model and compare the result with the existing model. $EmoH\_W$ and $Ext\_w$ show the comparison of the accuracy in figure 9. This result shows $EmoBang_{Hybrid}$ with Word2Vec has slightly better (0.05%) accuracy than the existing model with the Word2Vec technique. However, a significant contribution of our hybrid model is that we work with eight different emotional classes, and the existing model[20] works only with three emotional classes. **RQ5: LLM's performance on the dataset:** We conduct a small-scale extension that employs large language models (LLMs) in zero-shot and few-shot settings to align this study with recent advances in low-resource natural language processing (NLP). The zero-shot and few-shot paradigms leverage the intrinsic generalization capabilities of pre-trained transformer-based LLMs to perform classification with minimal or no task-specific training. In contrast, conventional supervised learning involves refining models on task-specific labeled datasets.

The model configuration is as follows: We evaluate two multilingual LLMs, BanglaBERT [8], and XLM-RoBERTa Large [3]. Using the prompt "Classify the emotion: [input text]", the models classify Bengali sentences into eight emotion categories under the zero-shot configuration without any fine-tuning. In the few-shot setting, we provide each model with ten labeled examples per class as in-context demonstrations prior to inference. With zero-shot accuracies between 84% and 86%, the results in Table 4 indicate that LLMs such as XLM-RoBERTa and BanglaBERT perform effectively without task-specific fine-tuning. Both models exhibit improved performance under few-shot conditions, achieving an F1-score of 91.74% and an accuracy of up to 91.22%. These results confirm that contemporary multilingual LLMs generalize well in low-resource language settings.

Table 4: Zero-shot and Few-shot Performance of LLMs

| Model | Setup | Accuracy (%) | Precision (%) | F1-score (%) |
|---|---|---|---|---|
| BanglaBERT | Zero-shot | 84.92 | 85.12 | 84.76 |
| BanglaBERT | Few-shot | 89.54 | 90.45 | 89.92 |
| XLM-RoBERTa Large | Zero-shot | 86.21 | 87.60 | 86.83 |
| XLM-RoBERTa Large | Few-shot | 91.22 | 92.35 | 91.74 |

The proposed ensemble model, *EmoBang$_{Ensemble}$*, outperforms conventional baselines and individual LLMs, achieving the highest performance (Accuracy = 93.69%, F1 = 93.67%). The model captures emotion-specific lexical and contextual dependencies in Bengali text through task-specific fine-tuning and AdaBoost-based integration of BERT embeddings.

In contrast, zero-shot and few-shot LLMs yield slightly lower accuracy due to their reliance on general-purpose representations that are not fully optimized for emotion detection. Hence, for low-resource languages, domain-optimized hybrid architectures such as *EmoBang$_{Ensemble}$* provide superior emotion classification performance, while LLMs remain a promising option for scalable, low-data applications.

## 5  Conclusion and Discussion

Although the experimental results are promising, it is important to recognize that the models' applicability in real-world scenarios may be limited. The models may not perform as effectively on text data from different sources or on other types of textual content. Furthermore, the accuracy scores achieved in this study may not accurately represent real-world performance, as the experiments are conducted on a specific dataset. Through comprehensive experimental evaluations, we demonstrate that our proposed models outperform existing approaches developed for Bengali texts in terms of accuracy, precision, and recall. We conduct an extensive empirical study that includes six baseline models and five feature engineering techniques, providing a thorough analysis of emotion detection in Bengali text. To the best of our knowledge, no prior research has investigated these different feature engineering techniques within the context of Bengali emotion detection. Among the two proposed models, the *EmoBang$_{ensemble}$* model achieves the highest accuracy of 93.69%.

Despite these encouraging results, our study has several limitations. The proposed EmoBang models have not yet been applied to datasets of varying sizes, structures, or linguistic domains. Additionally, they have not been tested on datasets from other languages. As part of our future research, we plan to apply the proposed models to larger and more diverse datasets drawn from multiple low-resource languages. This effort will allow us to evaluate the generalizability and scalability of our approach. We also intend to explore large language models (LLMs) in future work, as these models are trained on vast and diverse text corpora and may offer improved contextual understanding and robustness. Incorporating LLMs could potentially enhance emotion detection performance and broaden the applicability of our approach across multiple languages and domains.

**References**

1. Daraz website. https://www.daraz.com.bd/, accessed: 2023-05-24
2. Prothom alo newspaper website. https://www.prothomalo.com/, accessed: 202305-24
3. Xlm roberta large multilingual language model. https://dataloop.ai/library/model/, accessed: 2025-10-26
4. Alam, T., Khan, A., Alam, F.: Bangla text classification using transformers. arXiv preprint arXiv:2011.04446 (2020)
5. Alvi, N., Talukder, K.H.: Sentiment analysis of bengali text using countvectorizer with logistic regression. In: 2021 12th International Conference on Computing Communication and Networking Technologies (ICCCNT). pp. 01–05. IEEE (2021)
6. Azmin, S., Dhar, K.: Emotion detection from bangla text corpus using naive bayes classifier. In: 2019 4th International Conference on Electrical Information and Communication Technology (EICT). pp. 1–5. IEEE (2019)
7. Bashir, M.F., Javed, A.R., Arshad, M.U., Gadekallu, T.R., Shahzad, W., Beg, M.O.: Context-aware emotion detection from low-resource urdu language using deep neural network. ACM Trans. Asian Low-Resour. Lang. Inf. Process. **22**(5) (may 2023). https://doi.org/10.1145/3528576, https://doi.org/10.1145/3528576
8. Bhattacharjee, A., Hasan, T., Ahmad, W.U., Samin, K., Islam, M.S., Iqbal, A., Rahman, M.S., Shahriyar, R.: Banglabert: Language model pretraining and benchmarks for low-resource language understanding evaluation in bangla (2022), https://arxiv.org/abs/2101.00204
9. Bhowmik, N.R., Arifuzzaman, M., Mondal, M.R.H., Islam, M.: Bangla text sentiment analysis using supervised machine learning with extended lexicon dictionary. Natural Language Processing Research **1**(3-4), 34–45 (2021)
10. Binali, H., wu, C., Potdar, V.: Computational approaches for emotion detection in text. pp. 172 – 177 (05 2010). https://doi.org/10.1109/DEST.2010.5610650
11. Campan, A., Atnafu, T., Truta, T.M., Nolan, J.: Is data collection through twitter streaming api useful for academic research? In: 2018 IEEE international conference on big data (big data). pp. 3638–3643. IEEE (2018)
12. Das, B., Mandal, S., Mitra, P.: Bengali speech corpus for continuous auutomatic speech recognition system. In: 2011 International conference on speech database and assessments (Oriental COCOSDA). pp. 51–55. IEEE (2011)
13. Devlin, J., Chang, M.W., Lee, K., Toutanova, K.: Bert: Pre-training of deep bidirectional transformers for language understanding. arXiv preprint arXiv:1810.04805 (2018)
14. Diouf, R., Sarr, E.N., Sall, O., Birregah, B., Bousso, M., Mbaye, S.N.: Web scraping: state-of-the-art and areas of application. In: 2019 IEEE International Conference on Big Data (Big Data). pp. 6040–6042. IEEE (2019)
15. Gao, J., Ning, H., Han, Z., Kong, L., Qi, H.: Legal text classification model based on text statistical features and deep semantic features. In: FIRE (Working Notes). pp. 35–41 (2020)
16. Garg, K., Lobiyal, D.: Hindi emotionnet: A scalable emotion lexicon for sentiment classification of hindi text. ACM Transactions on Asian and Low-Resource Language Information Processing (TALLIP) **19**(4), 1–35 (2020)
17. Ghosh, T., Al Banna, M.H., Al Nahian, M.J., Uddin, M.N., Kaiser, M.S., Mahmud, M.: An attention-based hybrid architecture with explainability for depressive social media text detection in bangla. Expert Systems with Applications **213**, 119007 (2023)
18. Gomaa, W.H., Fahmy, A.A., et al.: A survey of text similarity approaches. international journal of Computer Applications **68**(13), 13–18 (2013)

19. Halibas, A.S., Shaffi, A.S., Mohamed, M.A.K.V.: Application of text classification and clustering of twitter data for business analytics. In: 2018 Majan international conference (MIC). pp. 1–7. IEEE (2018)
20. Hoq, M., Haque, P., Uddin, M.N.: Sentiment analysis of bangla language using deep learning approaches. In: Computing Science, Communication and Security: Second International Conference, COMS2 2021, Gujarat, India, February 6–7, 2021, Revised Selected Papers. pp. 140–151. Springer (2021)
21. Islam, T., Latif, S., Ahmed, N.: Using social networks to detect malicious bangla text content. In: 2019 1st International Conference on Advances in Science, Engineering and Robotics Technology (ICASERT). pp. 1–4. IEEE (2019)
22. Joulin, A., Grave, E., Bojanowski, P., Douze, M., Jégou, H., Mikolov, T.: Fasttext. zip: Compressing text classification models. arXiv preprint arXiv:1612.03651 (2016)
23. Liu, J.: Importance-smote: a synthetic minority oversampling method for noisy imbalanced data. Soft Computing **26**(3), 1141–1163 (2022)
24. Mikolov, T., Chen, K., Corrado, G., Dean, J.: Efficient estimation of word representations in vector space. arXiv preprint arXiv:1301.3781 (2013)
25. Mridha, M.F., Wadud, M.A.H., Hamid, M.A., Monowar, M.M., Abdullah-AlWadud, M., Alamri, A.: L-boost: Identifying offensive texts from social media post in bengali. Ieee Access **9**, 164681–164699 (2021)
26. Nikiforos, M.N., Deliveri, K., Kermanidis, K.L., Pateli, A.: Vocational domain identification with machine learning and natural language processing on wikipedia text: Error analysis and class balancing. Computers **12**(6), 111 (2023)
27. Parvin, T., Hoque, M.M.: An ensemble technique to classify multi-class textual emotion. Procedia Computer Science **193**, 72–81 (2021)
28. Purba, S.A., Tasnim, S., Jabin, M., Hossen, T., Hasan, M.K.: Document level emotion detection from bangla text using machine learning techniques. In: 2021 International Conference on Information and Communication Technology for Sustainable Development (ICICT4SD). pp. 406–411. IEEE (2021)
29. Pyreddy, S.R., Zaman, T.S.: Emoxpt: Analyzing emotional variances in human comments and llm-generated responses. In: 2025 IEEE 15th Annual Computing and Communication Workshop and Conference (CCWC). pp. 00088–00094 (2025). https://doi.org/10.1109/CCWC62904.2025.10903889
30. Rabeya, T., Ferdous, S., Ali, H.S., Chakraborty, N.R.: A survey on emotion detection: A lexicon based backtracking approach for detecting emotion from bengali text. In: 2017 20th international conference of computer and information technology (ICCIT). pp. 1–7. IEEE (2017)
31. Rabib, H.K., Galib, M., Nobo, T.M., Sathi, T.A., Islam, M.S., Kamal, A.R.M., Hossain, M.A.: Gender-based cyberbullying detection for under-resourced bangla language. In: 2022 12th International Conference on Electrical and Computer Engineering (ICECE). pp. 104–107. IEEE (2022)
32. Rahman, M., Seddiqui, M., et al.: Comparison of classical machine learning approaches on bangla textual emotion analysis. arXiv preprint arXiv:1907.07826 (2019)
33. Rahman, M.A., Kumar Dey, E.: Datasets for aspect-based sentiment analysis in bangla and its baseline evaluation. Data **3**(2), 15 (2018)
34. Rayhan, M.M., Al Musabe, T., Islam, M.A.: Multilabel emotion detection from bangla text using bigru and cnn-bilstm. In: 2020 23rd International Conference on Computer and Information Technology (ICCIT). pp. 1–6. IEEE (2020)

35. Ruposh, H.A., Hoque, M.M.: A computational approach of recognizing emotion from bengali texts. In: 2019 5th International Conference on Advances in Electrical Engineering (ICAEE). pp. 570–574. IEEE (2019)
36. Sadhu, J., Saha, M.R., Shahriyar, R.: An empirical study of gendered stereotypes in emotional attributes for bangla in multilingual large language models. arXiv preprint arXiv:2407.06432 (2024)
37. Sammons, M., Christodoulopoulos, C., Kordjamshidi, P., Khashabi, D., Srikumar, V., Roth, D.: Edison: Feature extraction for nlp, simplified. In: Proceedings of the Tenth International Conference on Language Resources and Evaluation (LREC'16). pp. 4085–4092 (2016)
38. Selva Birunda, S., Kanniga Devi, R.: A review on word embedding techniques for text classification. Innovative Data Communication Technologies and Application: Proceedings of ICIDCA 2020 pp. 267–281 (2021)
39. Shivhare, S.N., Khethawat, S.: Emotion detection from text. In: ArXiv. https://doi.org/10.48550/arXiv.1205.4944 (2012)
40. Shuhan, M.K.B., Dey, R., Saha, S., Anjum, M.S.U., Zaman, T.S.: A stylometric dataset for bengali poems. In: Proceedings of the 2022 6th International Conference on Natural Language Processing and Information Retrieval.
p. 176–180. NLPIR '22, Association for Computing Machinery, New York, NY, USA (2023). https://doi.org/10.1145/3582768.3582788, https://doi.org/10.1145/ 3582768.3582788
41. Sourav, M.S.U., Wang, H.: Transformer-based text classification on unified bangla multi-class emotion corpus. arXiv preprint arXiv:2210.06405 (2022)
42. Sreenidhi, M., Dhanya, S.S.S., Sahithi, R., Femi, P.S.: Human emotion recognition system using deep learning techniques. International Journal Of Engineering Research Technology (IJERT) **9**(07) (2020)
43. Tabassum, S.: Llm based sentiment classification from bangladesh e-commerce reviews. arXiv preprint arXiv:2510.01276 (2025)
44. Tripto, N.I., Ali, M.E.: Detecting multilabel sentiment and emotions from bangla youtube comments. In: 2018 International Conference on Bangla Speech and Language Processing (ICBSLP). pp. 1–6. IEEE (2018)
45. Tuhin, R.A., Paul, B.K., Nawrine, F., Akter, M., Das, A.K.: An automated system of sentiment analysis from bangla text using supervised learning techniques. In: 2019 IEEE 4th International Conference on Computer and Communication
Systems (ICCCS). pp. 360–364. IEEE (2019)